%% file: www.tex
\documentclass[sigconf]{acmart}
\usepackage{newcommand}
\usepackage{amsbsy}
\usepackage{multicol,graphicx,mathrsfs,pifont,amscd,latexsym,color,fancyhdr,url,longtable, pifont, subfigure,epstopdf,setspace,multirow,colortbl,tabularx, booktabs, bbm,listings, balance,color,indentfirst,algorithmic,algorithm, subfigure}
\usepackage{paralist}
\usepackage{amsmath}
\usepackage{stackengine}
\usepackage{relsize}
\usepackage{bbold}
\usepackage{amsfonts}
\usepackage{amssymb}
\usepackage{multirow}

\newcommand{\tabincell}[2]{\begin{tabular}{@{}#1@{}}#2\end{tabular}}

\usepackage{amsthm}
\newtheorem{theorem}{Theorem}
\theoremstyle{definition}
\newtheorem{defn}[theorem]{Definition}

\newcommand{\model}{Heterogeneous Graph Transformer}
\newcommand{\short}{HGT}
\newcommand{\sampling}{HGSampling}

\newcommand{\yd}[1]{\textit{{\color{red}#1 }}}
\newcommand{\zn}[1]{\textit{{\color{red}#1 }}}

\newcommand{\hide}[1]{} 
\newcommand{\vpara}[1]{\vspace{0.05in}\noindent\textbf{#1 }}

\copyrightyear{2020}
\acmYear{2020}
\acmConference[WWW '20]{Proceedings of The Web Conference 2020}{April 20--24, 2020}{Taipei, Taiwan}
\acmBooktitle{Proceedings of The Web Conference 2020 (WWW '20), April 20--24, 2020, Taipei, Taiwan}
\acmPrice{}
\acmDOI{10.1145/3366423.3380027}
\acmISBN{978-1-4503-7023-3/20/04}

\begin{document}

\title{Heterogeneous Graph Transformer}

\author{Ziniu Hu}
\authornote{This work was done when Ziniu was an intern at Microsoft Research.}
\affiliation{%
  \institution{University of California, Los Angeles}
}
\email{bull@cs.ucla.edu}

\author{Yuxiao Dong}
\affiliation{%
  \institution{Microsoft Research, Redmond}
}
\email{yuxdong@microsoft.com}

\author{Kuansan Wang}
\affiliation{%
  \institution{Microsoft Research, Redmond}
}
\email{kuansanw@microsoft.com}

\author{Yizhou Sun}
\affiliation{%
  \institution{University of California, Los Angeles}
}
\email{yzsun@cs.ucla.edu}

\begin{abstract}
\input{section/abstract}
\end{abstract}

\keywords{Graph Neural Networks; Heterogeneous Information Networks; Representation Learning; Graph Embedding; Graph Attention}

\maketitle

\section{Introduction}\label{sec:introduction}

\input{section/introduction}

\section{Preliminaries and Related Work}\label{sec:problem} \input{section/problem}

\section{Heterogeneous Graph Transformer}\label{sec:approach}
\input{section/approach}
\section{Web-scale \short\ Training}\label{sec:train}

\input{section/train}

\section{Evaluation}\label{sec:evaluation}
\input{section/evaluation}

\section{Conclusion}\label{sec:conclusion}
\input{section/conclusion}

\bibliographystyle{ACM-Reference-Format}\balance
\bibliography{www}

\end{document}

%% file: section/abstract.tex
Recent years have witnessed the emerging success of graph neural networks (GNNs) for modeling structured data. 
However, most GNNs are designed for homogeneous graphs, in which all nodes and edges belong to the same types, making them infeasible to represent  heterogeneous structures. 
In this paper, we present the \model\ (\short) architecture for modeling Web-scale heterogeneous graphs. 
To model heterogeneity, we design node- and edge-type dependent parameters to characterize the heterogeneous attention over each edge, empowering \short\ to maintain dedicated representations for different types of nodes and edges. 
To handle dynamic heterogeneous graphs, we introduce the relative temporal encoding technique into \short, which is able to capture the dynamic structural dependency with arbitrary durations. 
To handle Web-scale graph data, we design the heterogeneous mini-batch graph sampling algorithm---\sampling---for efficient and scalable training. 
Extensive experiments on the Open Academic Graph of 179 million nodes and 2 billion edges show that the proposed \short\ model consistently outperforms all the state-of-the-art GNN baselines by 9$\%$--21$\%$ on various downstream tasks. The dataset and source code of \short\ are publicly available at \url{https://github.com/acbull/pyHGT}.

\hide{
Heterogeneous and dynamic graphs are important abstractions for modeling relational data for many real-world systems, and modelling such graphs have been a challenging research topics. Recently, researchers have adopted deep learning into graph domain and achieve competitive performance in various graph-based applications. However, most existing GNN assume nodes are within same distribution and in static graphs, and thus cannot be directly adapted to heterogeneous and dynamic graphs. In this paper, we propose Heterogeneous Graph Transformer (HGT), for modeling web-scale heterogeneous and dynamic graphs. To model heterogeneity, we leverage the meta relation triplet of heterogeneous graph schema to define interaction and transform matrices, so that the model can capture both the common and specific patterns of different relationships using equal or even smaller parameters. To capture temporal information of graphs, we propose to keep a whole graph with edges happening in different time to avoid structure loss, and adopt relative temporal encoding to make the HGT capture dynamic dependency. We further design a heterogeneous graph sampling algorithm with an inductive timestamp assignment algorithm for training large-scale heterogeneous and dynamic graphs. We conduct experiments on Open Academic Graphs (OAG), which contains 0.1 billion nodes and 2 billion edges. The results show that our proposed HGT outperform all the state-of-the-art baselines on various downstream tasks.
}

%% file: section/introduction.tex
\begin{figure}[t!]
    \centering
    \includegraphics[width=0.48\textwidth, trim = 5 5 10 0, clip
    ] 
    {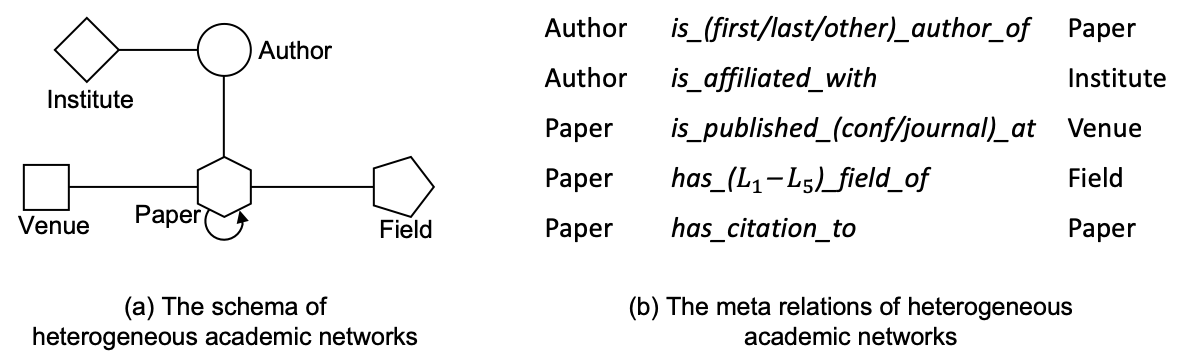}
    \caption{The schema and meta relations of Open Academic Graph (OAG). \textmd{Given a Web-scale heterogeneous graph, e.g., an academic network, \short\ takes only its one-hop edges as input without manually designing meta paths.}}
    \label{fig:schema}
\end{figure}

Heterogeneous graphs have been commonly used for abstracting and modeling complex systems, in which objects of different types interact with each other in various ways. 
Some prevalent instances of such systems include academic graphs, Facebook entity graph, LinkedIn economic graph, and broadly the Internet of Things network. 
For example, the  Open Academic Graph (OAG)~\cite{DBLP:conf/kdd/ZhangLTDYZGWSLW19} in Figure \ref{fig:schema} contains five types of nodes: papers, authors, institutions, venues (journal, conference, or preprint), and fields, as well as different types of relationships between them. 

Over the past decade, a significant line of research has been explored for mining heterogeneous graphs~\cite{Sun:BOOK2012}. 
One of the classical paradigms is to define and use meta paths to model heterogeneous structures, such as PathSim~\cite{Sun:VLDB11} and metapath2vec~\cite{dong2017metapath2vec}. 
Recently, in view of graph neural networks' (GNNs) success~\cite{gcn, graphsage, gat}, there are several attempts to adopt GNNs to learn with heterogeneous networks~\cite{DBLP:conf/esws/SchlichtkrullKB18,DBLP:conf/kdd/ZhangSHSC19,DBLP:conf/www/WangJSWYCY19,gt}. 
However, these works face several issues: 
First, most of them involve the design of meta paths  for each type of heterogeneous graphs, requiring specific domain knowledge; 
Second, they either simply assume that different types of nodes/edges share the same feature and representation space or keep distinct non-sharing weights for either node type or edge type alone, making them insufficient to capture heterogeneous graphs' properties; 
Third, most of them ignore the dynamic nature of every (heterogeneous) graph; 
Finally, their intrinsic design and implementation make them incapable of modeling Web-scale heterogeneous graphs.

Take OAG for example: First, the nodes and edges in OAG could have different feature distributions, e.g., papers have text features whereas institutions may have features from affiliated scholars, and coauthorships obviously differ from citation links; 
Second, OAG has been consistently evolving, e.g., 1) the volume of publications doubles every 12 years~\cite{dong2017century}, and 2) the KDD conference was more related to database in the 1990s whereas more to machine learning in recent years; 
Finally, OAG contains hundreds of millions of nodes and billions of relationships, leaving existing heterogeneous GNNs not scalable for handling it.

In light of these limitations and challenges, we propose to study heterogeneous graph neural networks with the goal of maintaining node- and edge-type dependent representations, capturing network dynamics, avoiding customized meta paths, and being scalable to Web-scale graphs. 
In this work, we present the \model\ (\short) architecture to deal with all these issues. 


To handle graph heterogeneity, we introduce the node- and edge-type dependent attention mechanism.
Instead of parameterizing each type of edges, the heterogeneous mutual attention in \short\ is defined by breaking down each edge $e=(s,t)$ based on its meta relation triplet, i.e., $\langle$ node type of $s$, edge type of $e$ between $s$ \& $t$, node type of $t \rangle$. Figure \ref{fig:schema} illustrates the meta relations of heterogeneous academic graphs. 
In specific, we use these meta relations to parameterize the weight matrices for calculating attention over each edge.
As a result, nodes and edges of different types are allowed to maintain their specific representation spaces.
Meanwhile, connected nodes in different types can still interact, pass, and aggregate messages without being restricted by their distribution gaps.
Due to the nature of its architecture, \short\ can incorporate information from high-order neighbors of different types through message passing across layers, which can be regarded as ``soft'' meta paths. 
That said, even if \short\ take only its one-hop edges as input without manually designing meta paths, the proposed attention mechanism can automatically and implicitly learn and extract ``meta paths'' that are important for different downstream tasks.

To handle graph dynamics, we enhance \short\ by proposing the relative temporal encoding (RTE) strategy. 
Instead of slicing the input graph into different timestamps, we propose to maintain all the edges happening in different times as a whole, and design the RTE strategy to model structural temporal dependencies with any duration length, and even with unseen and future timestamps. 
By end-to-end training, RTE enables \short\ to automatically learn the temporal dependency and evolution of heterogeneous graphs.

To handle Web-scale graph data, we design the first heterogeneous sub-graph sampling algorithm---\sampling---for mini-batch GNN training. 
Its main idea is to sample heterogeneous sub-graphs in which different types of nodes are with similar proportions, since the direct usage of existing (homogeneous) GNN sampling methods, such as GraphSage~\cite{graphsage}, FastGCN~\cite{fastgcn}, and LADIES~\cite{ladies}, results in highly imbalanced ones regarding to both node and edge types. 
In addition, it is also designed to keep the sampled sub-graphs dense for minimizing the loss of information. 
With \sampling, all the GNN models, including our proposed \short, can train and infer on arbitrary-size heterogeneous graphs. 

We demonstrate the effectiveness and efficiency of the proposed \model\ on the Web-scale Open Academic Graph comprised of 179 million nodes and 2 billion edges spanning from 1900 to 2019, making this the largest-scale
and longest-spanning representation learning yet performed on heterogeneous graphs. 
Additionally, we also examine it on domain-specific graphs: the computer science and medicine academic graphs. 
Experimental results suggest that \short\ can significantly improve various downstream tasks over state-of-the-art GNNs as well as dedicated heterogeneous models by 9--21$\%$. 
We further conduct case studies to show the proposed method can indeed  automatically capture the importance of implicit meta paths for different tasks.

%% file: section/problem.tex
In this section, we introduce the basic definition of heterogeneous graphs with network dynamics and review the recent development on graph neural networks (GNNs) and their heterogeneous variants. 
We also highlight the difference between \short \ and existing attempts on heterogeneous graph neural networks.

\subsection{Heterogeneous Graph Mining}
Heterogeneous graphs~\cite{Sun:BOOK2012} (a.k.a., heterogeneous information networks) are an important abstraction for modeling relational data for many real-world complex systems. Formally, it is defined as:
\theoremstyle{definition}
\begin{defn}{\textbf{Heterogeneous Graph:}}
A heterogeneous graph is defined as a directed graph $G = (\cV, \cE, \cA, \cR)$ where each node $v \in \cV$ and each edge $e \in \cE$ are associated with their type mapping functions $\tau(v): V \rightarrow \cA$ and $\phi(e): E \rightarrow \cR$, respectively. 
\end{defn}

\vpara{Meta Relation.}For an edge $e=(s, t)$ linked from source node $s$ to target node $t$, its {meta relation} is denoted as $\langle \tau(s), \phi(e), \tau(t) \rangle$. Naturally, $\phi(e)^{-1}$ represents the inverse of $\phi(e)$. The classical meta path paradigm~\cite{Sun:VLDB11,DBLP:conf/kdd/SunNHYYY12, Sun:BOOK2012} is defined as a sequence of such meta relation.

Notice that, to better model real-world heterogeneous networks, we assume that there may exist multiple types of relations between different types of nodes. 
For example, in OAG there are different types of relations between the \textit{author} and \textit{paper} nodes by considering the authorship order, i.e., ``the first author of'', ``the second author of'', and so on.

\vpara{Dynamic Heterogeneous Graph.}To model the dynamic nature of real-world (heterogeneous) graphs, we assign an edge $e=(s, t)$  a timestamp $T$, when node $s$ connects to node $t$ at $T$. 
If $s$ appears for the first time, $T$ is also assigned to $s$. 
$s$ can be associated with multiple timestamps if it builds connections over time. 

In other words, we assume that the timestamp of an edge is unchanged, denoting the time it is created. For example, when a paper published on a conference at time $T$, $T$ will be assigned to the edge between the paper and conference nodes. 
On the contrary, different timestamps can be assigned to a node accordingly. 
For example, the \textit{conference} node ``WWW'' can be assigned any year. $WWW@1994$ means that we are considering the first edition of WWW, which focuses more on internet protocol and Web infrastructure, while $WWW@2020$ means the upcoming WWW, which expands its research topics to social analysis, ubiquitous computing, search \& IR, privacy and society, etc. 

There have been significant lines of research on mining heterogenous graphs, such as node classification, clustering, ranking and representation learning~\cite{Sun:BOOK2012,Sun:VLDB11,DBLP:conf/kdd/SunNHYYY12,dong2017metapath2vec}, while the dynamic perspective of HGs has not been extensively explored and studied. 

\subsection{Graph Neural Networks}
Recent years have witnessed the success of graph neural networks for relational data~\cite{gcn,gat,graphsage}. 
Generally, a GNN can be regarded as using the input graph structure as the computation graph for message passing~\cite{DBLP:conf/icml/GilmerSRVD17}, during which the local neighborhood information is aggregated to get a more contextual representation. Formally, it has the following form:
\begin{defn}{\textbf{General GNN Framework:}}
Suppose $H^{l}[t]$ is the node representation of node $t$ at the $(l)$-th GNN layer, the update procedure from the $(l$-$1)$-th layer to the $(l)$-th layer is:
\begin{align}
H^{l}[t] \gets \underset{\forall s \in N(t), \forall e \in E(s,t)}{\textbf{Aggregate}}\bigg(  \textbf{Extract}\Big(H^{l-1}[s]; H^{l-1}[t], e\Big)\bigg)
\end{align}
where $N(t)$ denotes all the source nodes of node $t$ and $E(s,t)$  denotes all the edges from node $s$ to $t$. 
\end{defn}

The most important GNN operators are Extract($\cdot$) and Aggregate($\cdot$). 
Extract($\cdot$) represents the neighbor information extractor. 
It extract useful information from source node's representation $H^{l-1}[s]$, with the target node's representation $H^{l-1}[t]$ and the edge $e$ between the two nodes as query. 
Aggregate($\cdot$) gather the neighborhood information of souce nodes via some aggregation operators, such as \textit{mean, sum,} and \textit{max}, while more sophisticated pooling and normalization functions can be also designed.

Various (homogeneous) GNN architectures have been proposed following this framework.  
Kipf \textit{et al.}~\cite{gcn} propose graph convolutional network (GCN), which averages the one-hop neighbor of each node in the graph, followed by a linear projection and non-linear activation operations. 
Hamilton  \textit{et al.} propose GraphSAGE that generalizes GCN's aggregation operation from \textit{average} to \textit{sum, max} and a \textit{RNN unit}. 
Velickovi  \textit{et al.} propose graph attention network (GAT)~\cite{gat} by introducing the attention mechanism into GNNs, which allows GAT to assign different importance to nodes within the same neighborhood. 

\subsection{Heterogeneous GNNs}
Recently, studies have attempted to extend GNNs for modeling heterogeneous graphs. 
Schlichtkrull \textit{et al.}~\cite{DBLP:conf/esws/SchlichtkrullKB18} propose the relational graph convolutional networks (RGCN) to model knowledge graphs. 
RGCN keeps a distinct linear projection weight for each edge type. 
Zhang \textit{et al.}~\cite{DBLP:conf/kdd/ZhangSHSC19} present the heterogeneous graph neural networks (HetGNN) that adopts different RNNs for different node types to integrate multi-modal features. 
Wang \textit{et al.}~\cite{DBLP:conf/www/WangJSWYCY19} extend graph attention networks by maintaining different weights for different meta-path-defined edges. They also use high-level semantic attention to differentiate and aggregate information from different meta paths.

Though these methods have shown to be empirically better than the vanilla GCN and GAT models, they have not fully utilized the heterogeneous graphs' properties. All of them use either node type or edge type alone to determine GNN weight matrices. 
However, the node or edge counts of different types can vary greatly. 
For relations that don't have sufficient occurrences, it's hard to learn accurate relation-specific weights. To address this, we propose to consider parameter sharing for a better generalization. 
Given an edge $e=(s,t)$  with its meta relation as $\langle \tau(s), \phi(e), \tau(t) \rangle$, if we use three interaction matrices to model the three corresponding elements $\tau(s), \phi(e)$, and $\tau(t)$ in the meta relation, then the majority of weights could be shared. 
For example, in ``the first author of'' and ``the second author of'' relationships, their source and target node types are both \textit{author} to \textit{paper}, respectively. 
In other words, the knowledge about \textit{author} and \textit{paper} learned from one relation could be quickly transferred and adapted to the other one. 
Therefore, we integrate this idea with the powerful Transformer-like attention architecture, and propose \model. 

To summarize, the key differences between \short\ and existing attempts include:
\begin{enumerate}
    \item Instead of attending on node or edge type alone, we use the meta relation $\langle \tau(s), \phi(e), \tau(t) \rangle$ to decompose the interaction and transform matrices, enabling \short\ to capture both the common and specific patterns of different relationships using equal or even fewer parameters.
    \item Different from most of the existing works that are based on customized meta paths, we rely on the nature of the neural architecture to incorporate high-order heterogeneous neighbor information, which automatically learns the importance of implicit meta paths.  
    \item Most previous works don't take the dynamic nature of (heterogeneous) graphs into consideration, while we propose the relative temporal encoding technique to incorporate temporal information by using limited computational resources.
    \item None of the existing heterogeneous GNNs are designed for and experimented with Web-scale graphs, we therefore propose the heterogeneous Mini-Batch graph sampling algorithm designed for Web-scale graph training, enabling experiments on the billion-scale Open Academic Graph.
\end{enumerate}

%% file: section/approach.tex
\begin{figure*}[ht!]
    \centering
    \includegraphics[width=0.98\textwidth, trim = 10 0 10 0, clip
    ]{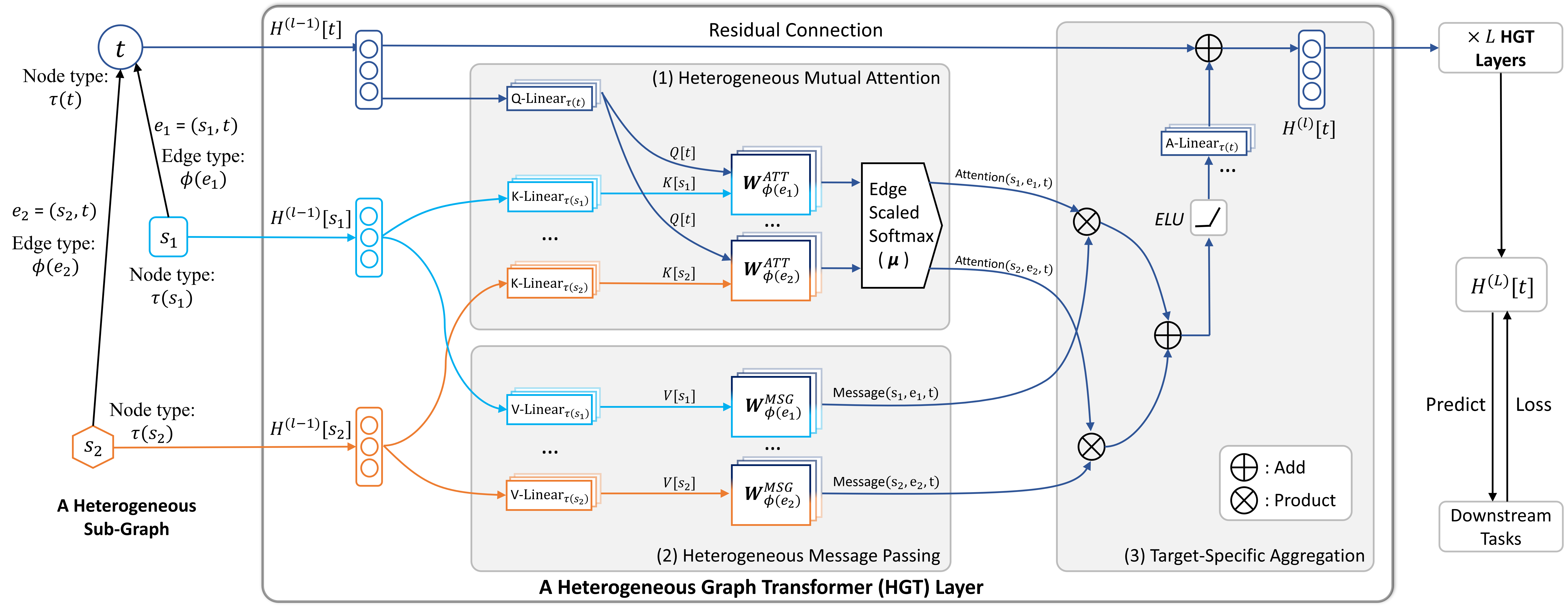}
    \caption{The Overall Architecture of Heterogeneous Graph Transformer. 
    \textmd{ Given a sampled heterogeneous sub-graph with $t$ as the target node, $s_1$ \& $s_2$ as source nodes, the \short\ model takes its edges $e_1=(s_1, t)$ \& $e_2=(s_2, t)$ and their corresponding meta relations $<\tau(s_1), \phi(e_1), \tau(t)>$ \& $<\tau(s_2), \phi(e_2), \tau(t)>$ as input to learn a  contextualized representation $H^{(L)}$ for each node, which can be used for downstream  tasks. 
    Color decodes the node type. 
    HGT includes three components: (1) meta relation-aware heterogeneous mutual attention,  (2) heterogeneous message passing from source nodes, and (3) target-specific heterogeneous message aggregation.}}
    \label{fig:my_label}
\end{figure*}



In this section, we present the \model\ (\short). 
Its  idea is to use the \textbf{meta relations} of heterogeneous graphs to parameterize weight matrices for the heterogeneous mutual attention, message passing, and propagation steps. 
To further incorporate network dynamics, we introduce a relative temporal encoding mechanism into the model.

\subsection{Overall \short\ Architecture}

Figure~\ref{fig:my_label} shows the overall architecture of \model. Given a sampled heterogeneous sub-graph (Cf. Section \ref{sec:train}), \short\ extracts all linked node pairs, where target node $t$ is linked by source node $s$ via edge $e$. The goal of \short\ is to aggregate information from source nodes to get a contextualized representation for target node $t$. Such process can be decomposed into three components: \textit{Heterogeneous Mutual Attention}, \textit{Heterogeneous Message Passing} and \textit{Target-Specific Aggregation}. 

We denote the output of the $(l)$-th \short\ layer as $H^{(l)}$, which is also the input of the $(l$+$1)$-th layer. 
By stacking $L$ layers, we can get the node representations of the whole graph $H^{(L)}$, which can be used for end-to-end training or fed into downstream tasks.  


\subsection{Heterogeneous Mutual Attention}

The first step is to calculate the mutual attention between source node $s$ and target node $t$. We first give a brief introduction to the general attention-based GNNs as follows: 
\begin{align}
H^{l}[t] \gets \underset{\forall s \in N(t), \forall e \in E(s,t)}{\textbf{Aggregate}}\Big(  \textbf{Attention}(s, t) \cdot \textbf{Message}(s)\Big)
\end{align}
where there are three basic operators: \textbf{Attention}, which estimates the importance of each source node; \textbf{Message}, which extracts the message by using only the source node $s$; and \textbf{Aggregate}, which aggregates the neighborhood message by the attention weight.

For example, the Graph Attention Network (GAT)~\cite{gat} adopts an additive mechanism as \textbf{Attention}, uses the same weight for calculating \textbf{Message}, and leverages the simple average followed by a nonlinear activation for the \textbf{Aggregate} step. 
Formally, GAT has
\begin{align}
    \textbf{Attention}_{GAT}(s, t) & = \underset{\forall s \in N(t)}{\text{Softmax}} \bigg(\Vec{a} \Big(WH^{l-1}[t] \mathbin\Vert WH^{l-1}[s]\Big)\bigg) \nonumber\\
    \textbf{Message}_{GAT}(s)  & = WH^{l-1}[s] \nonumber\\ 
    \textbf{Aggregate}_{GAT}(\cdot) & = \sigma \Big(\text{Mean}(\cdot)\Big) \nonumber
\end{align}
Though GAT is effective to give high attention values to important nodes, it assumes that $s$ and $t$ have the same feature distributions by using one weight matrix $W$. 
Such an assumption, as we've discussed in Section~\ref{sec:introduction}, 
is usually incorrect for heterogeneous graphs, where each type of nodes can have its own feature distribution.

In view of this limitation, we design the \textbf{Heterogeneous Mutual Attention} mechanism. 
Given a target node $t$, and all its neighbors $s \in N(t)$, which might belong to different distributions, we want to calculate their mutual attention grounded by their \textbf{meta relations}, i.e., the $\langle \tau(s), \phi(e), \tau(t) \rangle$ triplets. 

Inspired by the architecture design of Transformer~\cite{DBLP:conf/nips/VaswaniSPUJGKP17}, we  map target node $t$ into a Query vector, and source node $s$ into a Key vector, and calculate their dot product as attention. The key difference is that the vanilla Transformer uses a single set of projections for all words, while in our case each meta relation should have a distinct set of projection weights. To maximize parameter sharing while still maintaining the specific characteristics of different relations, we propose to parameterize the weight matrices of the interaction operators into a source node projection, an edge projection, and a target node projection. Specifically, we calculate the $h$-head {attention} for each edge $e=(s,t)$ (See Figure \ref{fig:my_label} (1)) by:
\begin{align}
\label{eq:hgt-att}
\textbf{Attention}_{HGT}(s,e,t)&  = \underset{\forall s \in N(t)}{\text{Softmax}}\Big(\underset{i \in [1,h]}{\mathlarger{\mathbin\Vert}}ATT\text{-}head^{i}(s,e,t)\Big)\\
ATT\text{-}head^{i}(s,e,t)&  = \Big(K^i(s)\ W^{ATT}_{\phi(e)}\ Q^i(t)^T\Big) \cdot \frac{{\mu}_{\langle \tau(s), \phi(e), \tau(t) \rangle}}{\sqrt{d}} \nonumber\\
K^i(s)&  = \text{K-Linear}^i_{\tau(s)}\Big({H}^{(l-1)}[s]\Big) \nonumber\\
Q^i(t)&  = \text{Q-Linear}^i_{\tau(t)}\Big(H^{(l-1)}[t]\Big) \nonumber
\end{align}
First, for the $i$-th attention head $ATT\text{-}head^{i}(s,e,t)$, we project the $\tau(s)$-type source node $s$ into the $i$-th \textit{Key} vector $K^i(s)$ with a linear projection K-Linear$^i_{\tau(s)}: \RR^{d} \rightarrow \RR^{\frac{d}{h}}$, where $h$ is the number of attention heads and $\frac{d}{h}$ is the vector dimension per head. 
Note that K-Linear$^i_{\tau(s)}$ is indexed by the source node $s$'s type $\tau(s)$, meaning that each type of nodes has a unique linear projection to maximally model the distribution differences. 
Similarly, we also project the target node $t$ with a linear projection Q-Linear$^i_{\tau(t)}$ into the $i-$th Query vector. 


Next, we need to calculate the similarity between the Query vector $Q^i(t)$ and Key vector $K^i(s)$. 
One unique characteristic of heterogeneous graphs is that there may exist different edge types (relations) between a node type pair, e.g., $\tau(s)$ and $\tau(t)$. 
Therefore, unlike the vanilla Transformer that directly calculates the dot product between the Query and Key vectors, we keep a distinct edge-based matrix $W^{ATT}_{\phi(e)}\in\RR^{\frac{d}{h}\times\frac{d}{h}}$ for each edge type $\phi(e)$. In doing so, the model can capture different semantic relations even between the same node type pairs. 
Moreover, since not all the relationships contribute equally to the target nodes, 
we add a prior tensor $\mu \in \RR^{|\cA|\times|\cR|\times|\cA|}$ to denote the general significance of each meta relation triplet, serving as an adaptive scaling to the attention. 

Finally, we concatenate $h$ attention heads together to get the attention vector for each node pair. 
Then, for each target node $t$, we gather all attention vectors from its neighbors $N(t)$ and conduct softmax, making it fulfill $\sum_{\forall s \in N(t)}\textbf{Attention}_{HGT}(s,e,t) = \mathbf{1}_{h \times 1}$.

\subsection{Heterogeneous Message Passing}
Parallel to the calculation of mutual attention, we pass information from source nodes to target nodes (See Figure \ref{fig:my_label} (2)). 
Similar to the attention process, we would like to incorporate the meta relations of edges into the message passing process to alleviate the distribution differences of nodes and edges of different types. For a pair of nodes $e =(s,t)$, we calculate its multi-head \textbf{Message} by:
\begin{align}
\textbf{Message}_{HGT}(s,e,t)&  = \underset{i \in [1,h]}{\mathlarger{\mathbin\Vert}}MSG\text{-}head^{i}(s,e, t)\\
MSG\text{-}head^{i}(s,e, t)&  = \text{M-Linear}^i_{\tau(s)}\Big({H}^{(l-1)}[s]\Big) \ W^{MSG}_{\phi(e)} \nonumber
\end{align}
To get the $i$-th message head $MSG\text{-}head^{i}(s,e,t)$, we first project the $\tau(s)$-type source node $s$ into the $i$-th message vector with a linear projection M-Linear$^i_{\tau(s)}: \RR^{d} \rightarrow \RR^{\frac{d}{h}}$. 
It is then followed by a matrix $W^{MSG}_{\phi(e)}\in\RR^{\frac{d}{h}\times\frac{d}{h}}$ for incorporating the edge dependency. 
The final step is to concat all $h$ message heads to get the  $\textbf{Message}_{HGT}(s,e, t)$ for each node pair.

\subsection{Target-Specific Aggregation}
With the heterogeneous multi-head attention and message calculated, we need to aggregate them from the source nodes to the target node (See Figure \ref{fig:my_label} (3)). 
Note that the softmax procedure in Eq. \ref{eq:hgt-att} has made the sum of each target node $t$'s attention vectors to one, we can thus simply use the attention vector as the weight to average the corresponding messages from the source nodes and get the updated vector $\widetilde{H}^{(l)}[t]$ as:
\begin{align}
\widetilde{H}^{(l)}[t] &= \underset{\forall s \in N(t)}{\mathlarger{\oplus}}\Big(\textbf{Attention}_{HGT}(s, e, t) \cdot \textbf{Message}_{HGT}(s, e, t)\Big). \nonumber
\end{align}
This aggregates information to the target node $t$ from all its neighbors (source nodes) of different feature distributions. 

The final step is to map target node $t$'s vector back to its type-specific distribution, indexed by its node type $\tau(t)$. 
To do so, we apply a linear projection A-Linear$_{\tau(t)}$ to the updated vector $\widetilde{H}^{(l)}[t]$, followed by residual connection~\cite{DBLP:conf/cvpr/HeZRS16} as:
\begin{align}
&H^{(l)}[t] = 
\text{A-Linear}_{\tau(t)}\Big(\sigma\big(\widetilde{H}^{(l)}[t]\big)\Big) + H^{(l-1)}[t]. \label{eq:output} 
\end{align}
In this way, we get the $l$-th \short\ layer's output $H^{(l)}[t]$ for the target node $t$. 
Due to the ``small-world'' property of real-world graphs, stacking the \short\ blocks for $L$ layers ($L$ being a small value) can enable each node reaching  a large proportion of nodes---with different types and relations---in the full graph. 
That is, \short\ generates a highly contextualized representation $H^{(L)}$ for each node, which can be fed into any models to conduct downstream heterogeneous network tasks, such as node classification and link prediction. 

Through the whole model architecture, we highly rely on using the \textbf{meta relation}---$\langle \tau(s), \phi(e), \tau(t) \rangle$---to parameterize the weight matrices separately. 
This can be interpreted as a trade-off between the model capacity and efficiency. Compared with the vanilla Transformer, our model distinguishes the operators for different relations and thus is more capable to handle the distribution differences in heterogeneous graphs. 
Compared with existing models that keep a distinct matrix for each meta relation as a whole, \short's triplet parameterization can better leverage the heterogeneous graph schema to achieve parameter sharing. 
On one hand, relations with few occurrences can benefit from such parameter sharing for fast adaptation and generalization. 
On the other hand, different relationships' operators can still maintain their specific characteristics by using a much smaller parameter set.

\begin{figure}[t!]
    \centering
    \includegraphics[width=0.47\textwidth, trim = 10 0 10 0, clip
    ]{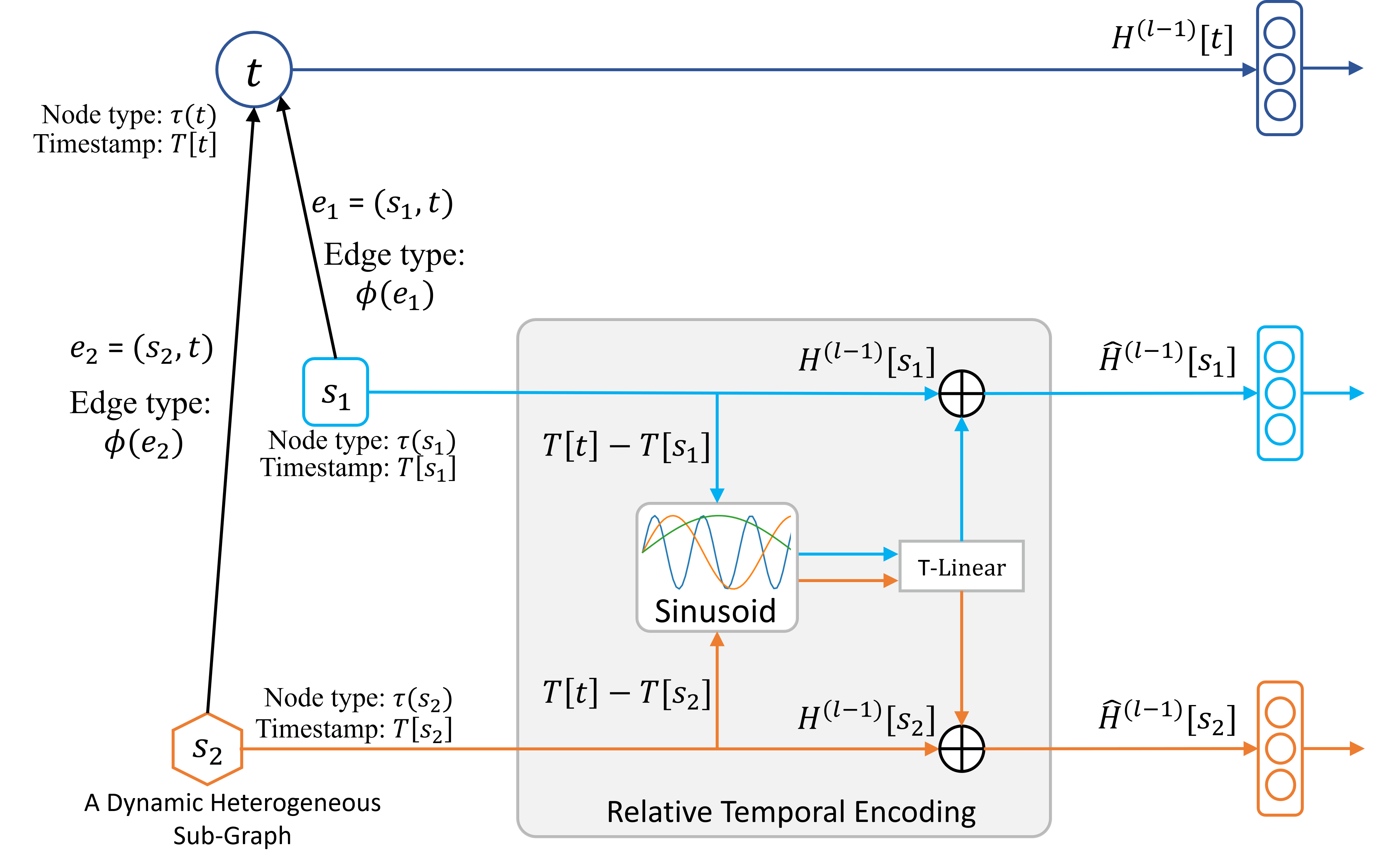}
    \caption{Relative Temporal Encoding (RTE) to model graph dynamic. \textmd{Nodes  are associated with timestamps $T(\cdot)$. After the RTE process, the temporal augmented representations are fed to the \short\ model.}}
    \label{fig:rte}
\end{figure}

\subsection{Relative Temporal Encoding}
By far, we present \short---a graph neural network for modeling heterogeneous graphs. 
Next, we introduce the Relative Temporal Encoding (RTE) technique for \short\ to handle graph dynamic.

The traditional way to incorporate temporal information is to construct a separate graph for each time slot. 
However, such a procedure may lose a large portion of structural dependencies across different time slots. 
Meanwhile, the representation of a node at time $t$ might rely on edges that happen at other time slots. 
Therefore, 
a proper way to model dynamic graphs is to maintain all the edges happening at different times and allow nodes and edges with different timestamps to interact with each other.

In light of this, we propose the Relative Temporal Encoding (RTE) mechanism to model the dynamic dependencies in heterogeneous graphs. 
RTE is inspired by Transformer's positional encoding method~\cite{DBLP:conf/nips/VaswaniSPUJGKP17, DBLP:conf/naacl/ShawUV18}, which has been shown successful to capture the sequential dependencies of words in long texts. 

Specifically, given a source node $s$ and a target node $t$, along with their corresponding timestamps $T(s)$ and $T(t)$, we denote the relative time gap $\Delta T(t,s) = T(t) - T(s)$ as an index to get a relative temporal encoding $RTE(\Delta T(t,s))$. 
Noted that the training dataset will not cover all possible time gaps, and thus  $RTE$ should be capable of generalizing to unseen times and time gaps. 
Therefore, we adopt a fixed set of sinusoid functions as basis, with a tunable linear projection T-Linear\footnote{For simplicity, we denote a linear projection L $:\RR^{a}\rightarrow \RR^{b}$ as a function to conduct linear transformation to vector $x\in\RR^{a}$ as: L$(x)=Wx+b$, where matrix $W\in\RR^{a+b}$ and bias $b\in\RR^{b}$. $W$ and $b$ are learnable parameters for L.}$: \RR^{d} \rightarrow \RR^{d}$ as $RTE$:
\begin{align}
   Base\big(\Delta T(t,s), 2i\big) & = sin\Big(\Delta T_{t,s} / 10000^{\frac{2i}{d}}\Big)\\ 
   Base\big(\Delta T(t,s), 2i+1\big) & = cos\Big(\Delta T_{t,s} / 10000^{\frac{2i+1}{d}}\Big)\\ 
   RTE\big(\Delta T(t,s)\big) & = \text{T-Linear}\Big( Base(\Delta T_{t,s}) \Big)
\end{align}
Finally, the temporal encoding relative to the target node $t$ is added to the source node $s$' representation as follows:
\begin{align}
    \widehat{H}^{(l-1)}[s] = H^{(l-1)}[s] + RTE\big(\Delta T(t,s)\big)
\end{align}
In this way, the temporal augmented representation $\widehat{H}^{(l-1)}$ will capture the relative temporal information of source node $s$ and target node $t$. The RTE procedure is illustrated in the Figure \ref{fig:rte}. 

%% file: section/train.tex
 \begin{figure*}[ht!]
    \centering
    \includegraphics[width=1.0\textwidth]{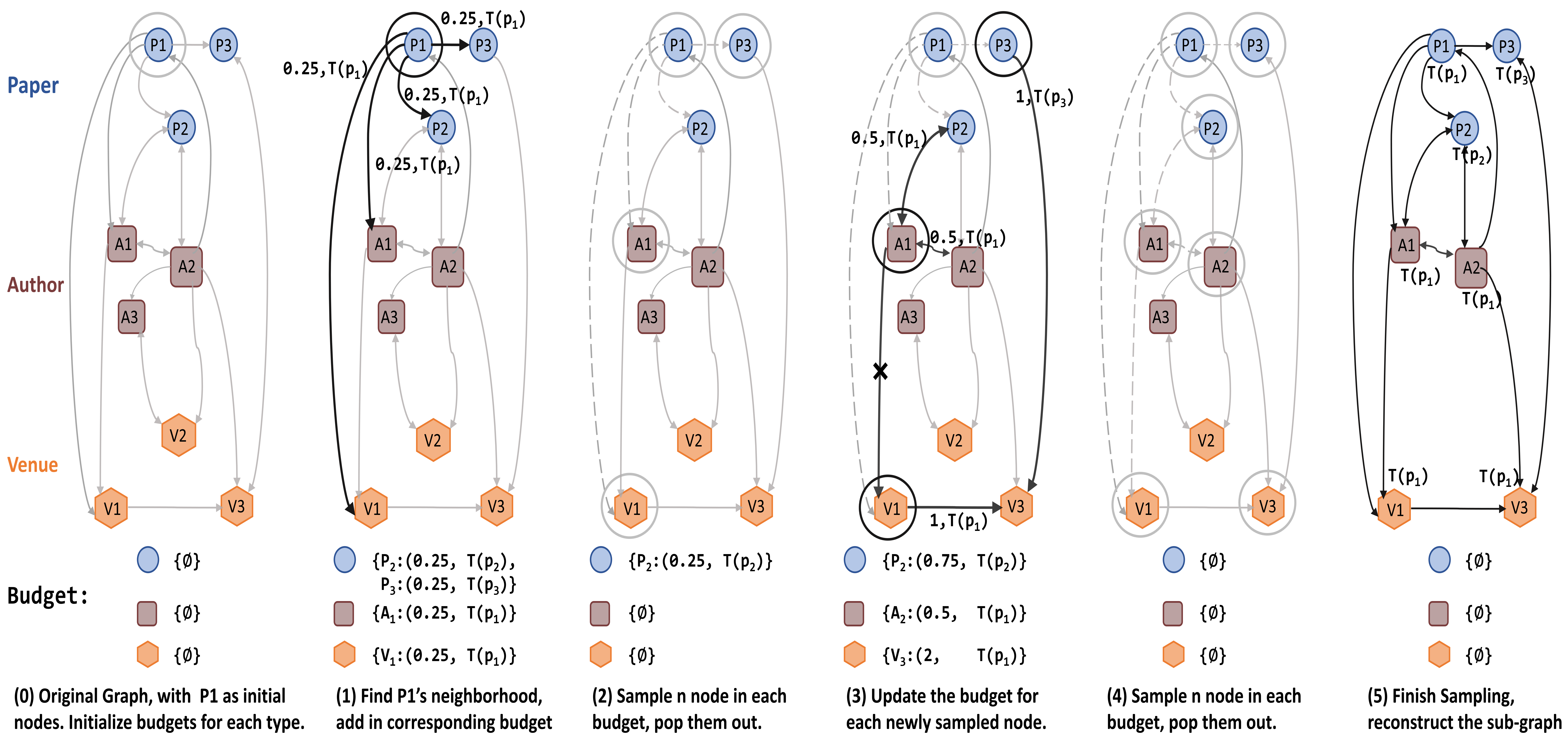}
    \caption{\sampling\ with Inductive Timestamp Assignment.}
    \label{fig:sample}
\end{figure*}

In this section, we present \short's strategies for training Web-scale heterogeneous graphs with dynamic information, including an efficient Heterogeneous Mini-Batch Graph Sampling algorithm---\sampling---and an inductive timestamp assignment method. 

\subsection{\sampling}

The full-batch GNN~\cite{gcn} training requires the calculation of all node representations per layer, 
making it not scalable for Web-scale graphs. 
To address this issue, different sampling-based methods~\cite{graphsage,fastgcn,DBLP:conf/icml/ChenZS18,ladies} have been proposed to train GNNs on a subset of nodes. 
However, directly using them for heterogeneous graphs is prone to get sub-graphs that are extremely imbalanced regarding different node types, due to that the degree distribution and the total number of nodes for each type can vary dramatically. 

To address this issue, we propose an efficient Heterogeneous Mini-Batch Graph Sampling algorithm---\sampling---to enable both \short\ and traditional GNNs to handle Web-scale heterogeneous graphs. 
\sampling\ is able to 1) keep a similar number of nodes and edges for each type and 2) keep the sampled sub-graph dense to minimize the information loss and reduce the sample variance.

Algorithm \ref{alg:sample} outlines the \sampling\ algorithm. 
Its basic idea is to keep a separate node budget $B[\tau]$ for each node type $\tau$ and to sample an equal number of nodes per type with an importance sampling strategy to reduce variance. 
Given node $t$ already sampled, we add all its direct neighbors into the corresponding budget with Algorithm~\ref{alg:budget}, and add $t$'s normalized degree to these neighbors in line~\ref{line:update}, which will then be used to calculate the sampling probability. 
Such normalization is equivalent to accumulate the random walk probability of each sampled node to its neighborhood, avoiding the sampling being dominated by high-degree nodes. 
Intuitively, the higher such value is, the more a candidate node is correlated with the currently sampled nodes, and thus should be given a higher probability to be sampled.

After the budget is updated, we then calculate the sampling probability in Algorithm~\ref{alg:sample} line~\ref{line:prob}, where we calculate the square of the cumulative normalized degree of each node $s$ in each budget. 
As proved in~\cite{ladies}, using such sampling probability can reduce the sampling variance. Then, we sample $n$ nodes in  type $\tau$ by using the calculated probability, add them into the output node set, update its neighborhood to the budget, and remove it out of the budget in lines~\ref{line:samp}--\ref{line:end}. 
Repeating such procedure for $L$ times, we get a sampled sub-graph with $L$ depth from the initial nodes. Finally, we reconstruct the adjacency matrix among the sampled nodes. 
By using the above algorithm, the sampled sub-graph contains a similar number of nodes per  type (based on the separate node budget), and is sufficiently dense to reduce the sampling variance (based on the normalized degree and importance sampling), making it suitable for training GNNs on Web-scale heterogeneous graphs.

\begin{algorithm}[tb] 
\caption{Heterogeneous Mini-Batch Graph Sampling} 
\label{alg:sample} 
\begin{algorithmic}[1] 
\REQUIRE
Adjacency matrix $A$ for each $\langle \tau(s), \phi(e), \tau(t) \rangle$ relation pair; Output node Set $OS$; Sample number $n$ per node type; Sample depth $L$.\\
\ENSURE
Sampled node set $NS$; Sampled adjacency matrix $\hat{A}$.
\STATE  $NS \gets OS$ // Initialize sampled node set as output node set.
\STATE  Initialize an empty Budget $B$ storing nodes for each node type with normalized degree. 
\FOR {$t \in NS$}
    \STATE  Add-In-Budget($B$, $t$, $A$, $NS$) // Add neighbors of $t$ to $B$.
\ENDFOR
\FOR {$l \gets 1$ to $L$}
    \FOR {source node type $\tau \in B$}
        \FOR {source node $s \in B[\tau]$}
            \STATE  $\ prob^{(l-1)}[\tau][s] \gets \frac{B[\tau][s]^2}{\|B[\tau]\|_2^2}$ // Calculate sampling probability for each source node $s$ of node type $\tau$. \label{line:prob}
        \ENDFOR
        \STATE  Sample $n$ nodes ${\{t_i\}}_{i=1}^n$ from $B[\tau]$ using $prob^{(l-1)}[\tau]$.
        \FOR {$t \in {\{t_i\}}_{i=1}^n$} \label{line:samp}
            \STATE  $OS[\tau].add(t)$ // Add node $t$ into Output node set.
            \STATE  Add-In-Budget($B$, $t$, $A$, $NS$) // Add neighbors of $t$ to $B$.
            \STATE  $B[\tau].pop(t)$ // Remove sampled node $t$ from Budget. \label{line:end}
        \ENDFOR
    \ENDFOR
\ENDFOR
\STATE  Reconstruct the sampled adjacency matrix $\hat{A}$ among the sampled nodes $OS$ from $A$.
\RETURN $OS$ and $\hat{A}$; 
\end{algorithmic} 
\end{algorithm}

\begin{algorithm}[tb] 
\caption{Add-In-Budget} 
\label{alg:budget} 
\begin{algorithmic}[1] 
\REQUIRE
Budget $B$ storing nodes for each type with normalized degree; Added node $t$; Adjacency matrix $A$ for each $\langle \tau(s), \phi(e), \tau(t) \rangle$ relation pair; Sampled node set $NS$.\\
\ENSURE
Updated Budget $B$.
     \FOR {each possible source node type $\tau$ and edge type $\phi$}
        \STATE  $\hat{D}_t \gets 1 \ /\ len\Big(A_{\langle \tau, \phi, \tau(t) \rangle}[t]\Big)$ // get normalized degree of added node $t$ regarding to $\langle \tau, \phi, \tau(t) \rangle$.
        \FOR {source node $s$ in $A_{\langle \tau, \phi, \tau(t) \rangle}[t]$}
            \IF {$s$ has not been sampled ($s \not\in NS$)} \label{line:check}
                 \IF {$s$ has no timestamp}
                     \STATE  $s.time = t.time$ // Inductively inherit timestamp. \label{line:time}
                 \ENDIF
                \STATE  $B[\tau][s] \gets B[\tau][s] + \hat{D}_t$\ \ \ // Add candidate node $s$ to budget $B$ with target node $t$'s normalized degree. \label{line:update}
            \ENDIF
        \ENDFOR  
    \ENDFOR
\RETURN Updated Budget $B$
\end{algorithmic} 
\end{algorithm}

\subsection{ Inductive Timestamp Assignment}

Till now we have assumed that each node $t$ is assigned with a timestamp $T(t)$. However, in real-world heterogeneous graphs, many nodes are not associated with a fixed time. Therefore, we need to assign different timestamps to it. We denote these nodes as {plain nodes}. For example, the WWW conference is held in both 1974 and 2019, and the WWW node in these two years has dramatically different research topics. Consequently, we need to decide which timestamp(s) to attach to the WWW node. 

There also exist {event nodes} in heterogeneous graphs that have an explicit timestamp associated with them. 
For example, the paper node should be associated with its publication behavior and therefore attached to its publication date. 

We propose an inductive timestamp assignment algorithm to assign plain nodes timestamps based on event nodes that they are linked with. 
The algorithm is shown in Algorithm~\ref{alg:budget} line~\ref{line:time}. 
The  idea is that plan nodes inherit the timestamps from event nodes. 
We examine whether the candidate source node is an event node. If yes, like a paper published at a specific year, we keep its timestamp for capturing temporal dependency. 
If no, like a conference that can be associated with any timestamp, we inductively assign the associated node's timestamp, such as the published year of its paper, to this plain node. In this way, we can adaptively assign timestamps during the sub-graph sampling procedure.



%% file: section/evaluation.tex

\begin{table*}[th]
\centering
\footnotesize
\begin{tabular}{c|rr|rrrrr|rrrrr} 
\toprule
Dataset & $\#$nodes & $\#$edges & $\#$papers & $\#$authors & $\#$fields & $\#$venues & $\#$institutes & $\#$P-A & $\#$P-F & $\#$P-V & $\#$A-I & $\#$P-P \\ 
\midrule
CS & 11,732,027 & 107,263,811 & 5,597,605 & 5,985,759 &  119,537&  27,433 & 16,931   & 15,571,614 & 47,462,559 & 5,597,606 & 7,190,480 & 31,441,552\\ 
\midrule
Med & 51,044,324 & 451,468,375 & 21,931,587 & 28,779,507&  289,930 &  25,044&  18,256  &85,620,479 & 149,728,483&21,931,588 & 28,779,507& 165,408,318\\ 
\midrule
OAG & 178,663,927 & 2,236,196,802  & 89,606,257 & 88,364,081 &  615,228&  53,073&  25,288 & 300,853,688&  657,049,405&  89,606,258&  167,449,933 & 1,021,237,518\\
\bottomrule
\end{tabular}
\caption{Open Academic Graph (OAG) Statistics.} 
\label{tab:stat} 
\end{table*}

In this section, we evaluate the proposed \model\ on three heterogeneous academic graph datasets. 
We conduct the Paper-Field prediction, Paper-Venue prediction, and Author Disambiguation tasks. 
We also take case studies to demonstrate how \short\ can automatically learn and extract meta paths that are important for downstream tasks\footnote{The dataset and code are publicly available at \url{https://github.com/acbull/pyHGT}.}.

\subsection{Web-Scale Datasets}

To examine the performance of the proposed model and its real-world applications, we use the Open Academic Graph (OAG)~\cite{DBLP:conf/www/SinhaSSMEHW15,tang2008arnetminer,DBLP:conf/kdd/ZhangLTDYZGWSLW19} as our experimental basis. 
OAG consists of more than 178 million nodes and 2.236 billion edges---the largest publicly available heterogeneous academic dataset. 
In addition, all papers in OAG are associated with their publication dates, spanning from 1900 to 2019. 

To test the generalization of the proposed model, we also construct two domain-specific subgraphs from OAG: the Computer Science (CS) and Medicine (Med) academic graphs. 
The graph statistics are listed in Table \ref{tab:stat}, in which P--A, P--F, P--V, A--I, and P--P denote the edges between paper and author, paper and field, paper and venue, author and institute, and the citation links between two papers.

Both the CS and Med graphs contain tens of millions of nodes and hundreds of millions of edges, making them at least one magnitude larger than the other CS (e.g., DBLP) and Med (e.g., Pubmed) academic datasets that are commonly used in existing heterogeneous GNN and heterogeneous graph mining studies. 
Moreover, the three datasets used are far more distinguishable than previously wide-adopted small citation graphs used in GNN studies, such as Cora, Citeseer, and Pubmed~\cite{gcn,gat}, which only contain thousands of nodes.

There are totally five node types: `Paper', `Author', `Field', `Venue', and `Institute'. 
The `Field' nodes in OAG are categorized into six levels from $L_0$ to $L_5$, which are organized with a hierarchical tree. 
Therefore, we differentiate the `Paper--Field' edges corresponding to the field level. 

In addition, we differentiate the different author orders (i.e., the first author, the last one, and others) and venue types (i.e., journal, conference, and preprint) as well. 
Finally, the `Self' type corresponds to the self-loop connection, which is widely added in GNN architectures. 
Except the `Self' relationship, which are symmetric, all other relation types $\phi$ have a reverse relation type $\phi^{-1}$.

\subsection{Experimental Setup}


\vpara{Tasks and Evaluation.}
We evaluate the \short\ model on four different real-world downstream tasks: the prediction of Paper--Field ($L_1$), Paper--Field ($L_2$), and Paper--Venue, and Author Disambiguation. 
The goal of the first three node classification tasks is to predict the correct $L_1$ and $L_2$ fields that each paper belongs to or the venue it is published at, respectively. 
We use different GNNs to get the contextual node representation of the paper and use a softmax output layer to get its classification label. 
For author disambiguation, we select all the authors with the same name and their associated papers. 
The task is to conduct link prediction between these papers and candidate authors. 
After getting the paper and author node representations from GNNs, we use a Neural Tensor Network to get the probability of each author-paper pair to be linked.

For all tasks, we use papers published before the year 2015 as the training set, papers between 2015 and 2016 for validation, and papers between 2016 and 2019 as testing. 
We choose NDCG and MRR, which are two widely adopted ranking metrics~\cite{DBLP:books/daglib/0027504, DBLP:series/synthesis/2014Li}, as the evaluation metrics. 
All models are trained for 5 times and, the mean and standard variance of test performance are reported.

\vpara{Baselines.}We compare \short\ with two classes of state-of-art graph neural networks. 
All baselines as well as our own model, 
are implemented via the PyTorch Geometric (PyG) package~\cite{pyG}. 

 The first class of GNN baselines is designed for homogeneous graphs, including:
\begin{itemize}
    \item Graph Convolutional Networks (GCN)~\cite{gcn}, which simply averages the neighbor's embedding followed by linear projection. We use the implementation provided in PyG.
    \item Graph Attention Networks (GAT)~\cite{gat}, which adopts multi-head additive attention on neighbors. We use the implementation provided in PyG. 
\end{itemize}

The second class considered is several dedicated heterogeneous GNNs as baselines,  including: 
\begin{itemize}
    \item Relational Graph Convolutional Networks (RGCN)~\cite{DBLP:conf/esws/SchlichtkrullKB18}, which keeps a different weight for each relationship, i.e., a relation triplet. We use the implementation provided in PyG.
    \item Heterogeneous Graph Neural Networks (HetGNN)~\cite{DBLP:conf/kdd/ZhangSHSC19}, which adopts different Bi-LSTMs for different node type for aggregating neighbor information. We re-implement this model in PyG following the authors' official code.
    \item Heterogeneous Graph Attention Networks (HAN)~\cite{DBLP:conf/www/WangJSWYCY19} design hierarchical attentions to aggregate neighbor information via different meta paths. We re-implement this model in PyG following the authors' official code.
\end{itemize}

In addition, to systematically analyze the effectiveness of the two major components of \short, i.e., Heterogeneous weight parameterization (Heter) and Relative Temporal Encoding (RTE), we conduct an ablation study, but comparing with models that remove these components. Specifically, we use $-Heter$ to denote models that uses the same set of weights for all meta relations, and use $-RTE$ to denote models that doesn't include relative temporal encoding. By considering all the permutations, we have: \short$_{-Heter}^{-RTE}$, \short$_{-Heter}^{+RTE}$, \short$_{+Heter}^{-RTE}$ and \short$_{+Heter}^{+RTE}$\footnote{Unless other stated, \short\ refers to \short$_{+Heter}^{+RTE}$.}.

We use our \sampling\ algorithm proposed in Section~\ref{sec:train} for all baseline GNNs to handle the large-scale OAG graph. To avoid data leakage, we remove out the links we aim to predict (e.g., the Paper-Field link as the label) from the sub-graph.

\begin{table*}[!tp]
\centering
\small
\renewcommand\arraystretch{1.3}
\setlength{\tabcolsep}{3pt}
\begin{tabular}{c|c|c|ccccc|cccc} 
\toprule
\multicolumn{3}{c|}{GNN Models} & GCN~\cite{gcn}  & RGCN~\cite{DBLP:conf/esws/SchlichtkrullKB18}& GAT~\cite{gat}& HetGNN~\cite{DBLP:conf/kdd/ZhangSHSC19} & HAN~\cite{DBLP:conf/www/WangJSWYCY19} &{ \short$_{-Heter}^{-RTE}$} & \short$_{-Heter}^{+RTE}$ & \short$_{+Heter}^{-RTE}$ & \short$_{+Heter}^{+RTE}$ \\ \midrule

\multicolumn{3}{c|}{$\#$ of Parameters} &1.69M   & 8.80M   & 1.69M   & 8.41M   & 9.45M   &3.12M   &3.88M & 7.44M   & 8.20M \\
\midrule
\multicolumn{3}{c|}{Batch Time} &  0.46s  &   1.24s &   0.97s &   1.35s &   2.27s &  1.11s &  1.14s &   1.48s &  1.50s \\
\midrule
    \multirow{10}{*}{\tabincell{c}{CS}} 
        & \multirow{2}{*}{Paper--Field ($L_1$)} & NDCG 
         &.608$\pm$.062 & .603$\pm$.065 & .622$\pm$.071 & .612$\pm$.063 &.618$\pm$.058 & .662$\pm$.051 & .689$\pm$.042 &.705$\pm$.036 & \textbf{.718$\pm$.014}\\ 
         ~&~& MRR 
         &.679$\pm$.069 & .683$\pm$.056 & .694$\pm$.065 & .689$\pm$.060 &.691$\pm$.051 & .751$\pm$.036& .779$\pm$.027 &.799$\pm$.023 & \textbf{.823$\pm$.019}\\
    \cmidrule{2-12}
        ~ & \multirow{2}{*}{Paper--Field ($L_2$)} & NDCG & .344$\pm$.021 & .322$\pm$.053& .357$\pm$.058& .346$\pm$.071 &.352$\pm$.051& .362$\pm$.048& .371$\pm$.043 & .379$\pm$.047&\textbf{.403$\pm$.041}  \\
        ~&~& MRR &.353$\pm$.053 &.340$\pm$.061 &.382$\pm$.057 & .373$\pm$.051 &.388$\pm$.065 & .394$\pm$.072&.397$\pm$.064&
        .414$\pm$.076& \textbf{.439$\pm$.078}\\
    \cmidrule{2-12}
        ~ & \multirow{2}{*}{\tabincell{c}{Paper--Venue}} & NDCG  &.406$\pm$.081 & .412$\pm$.076  & .437$\pm$.082 & .431$\pm$.074&  .449$\pm$.072 & .456$\pm$.069&.461$\pm$.066 &.468$\pm$.074&  \textbf{.473$\pm$.054}  \\  ~&~& MRR & .215$\pm$.066 &.216$\pm$.105&.239$\pm$.089 & .245$\pm$.069& .254$\pm$.074 & .258$\pm$.085& .265$\pm$.090 &.275$\pm$.089 &\textbf{.288$\pm$.088}\\
    \cmidrule{2-12}
        ~ & \multirow{2}{*}{\tabincell{c}{Author\\Disambiguation}} & NDCG &.826$\pm$.039 & .835$\pm$.042&.864$\pm$.051 & .850$\pm$.056& .859$\pm$.053 & .867$\pm$.048 &.875$\pm$.046 & .886$\pm$.048& \textbf{.894$\pm$.034}\\  ~&~& MRR &.661$\pm$.045&.665$\pm$.054 &.694$\pm$.052 & .668$\pm$.061 &.688$\pm$.049 & .703$\pm$.036& .712$\pm$.032 &.727$\pm$.038 & \textbf{.732$\pm$.038}  \\
\midrule
\multirow{10}{*}{\tabincell{c}{Med}} 
          & \multirow{2}{*}{Paper--Field ($L_1$)} 
          & NDCG &.560$\pm$.056 & .571$\pm$.061 & .584$\pm$.076 & .598$\pm$.068 & .607$\pm$.054&.654$\pm$.048 & .667$\pm$.045 &.683$\pm$.037 & \textbf{.709$\pm$.029}  \\  
          ~&~& MRR &.465$\pm$.055 &.470$\pm$.082 &.493$\pm$.069 & .509$\pm$.054& .575$\pm$.057 &.620$\pm$.066 &.642$\pm$.062 &.659$\pm$.055 & \textbf{.688$\pm$.048}\\
    \cmidrule{2-12}
        ~ & \multirow{2}{*}{Paper--Field ($L_2$)} & NDCG& .334$\pm$.035 & .337$\pm$.051& .344$\pm$.063&.342$\pm$.048  &.350$\pm$.059 & .359$\pm$.053& .365$\pm$.047 & .374$\pm$.050&\textbf{.384$\pm$.046}\\
        ~&~& MRR &.337$\pm$.061 &.343$\pm$.063 &.370$\pm$.058 &.373$\pm$.061 &.379$\pm$.052 & .385$\pm$.071& .397$\pm$.069&
        .408$\pm$.071& \textbf{.417$\pm$.074}\\
    \cmidrule{2-12}
        ~ & \multirow{2}{*}{\tabincell{c}{Paper--Venue }} & NDCG  &.377$\pm$.059
         & .383$\pm$.062 &.388$\pm$.065&  .412$\pm$.057& .416$\pm$.068& .421$\pm$.083 & .432$\pm$.078  &\textbf{.446$\pm$.083} &.445$\pm$.085  \\  ~&~& MRR &.211$\pm$.045 & .217$\pm$.058 &.244$\pm$.091 & .259$\pm$.072& .271$\pm$.056 & .277$\pm$.081&.282$\pm$.085 &.288$\pm$.074 & \textbf{.291$\pm$.062}\\
    \cmidrule{2-12}
        ~ & \multirow{2}{*}{\tabincell{c}{Author\\Disambiguation}}&MRR &.776$\pm$.042 & .779$\pm$.048&.828$\pm$.044 &.824$\pm$.058 & .834$\pm$.056 & .838$\pm$.047&.844$\pm$.041 & .864$\pm$.043& \textbf{.871$\pm$.040}\\  ~&~& NDCG &.614$\pm$.051&.625$\pm$.049 &.663$\pm$.046 & .659$\pm$.061 &.667$\pm$.053 &.683$\pm$.055 &.691$\pm$.046 &.708$\pm$.041 & \textbf{.718$\pm$.043}  \\ 
\midrule
\multirow{10}{*}{\tabincell{c}{OAG}} 
          & \multirow{2}{*}{Paper--Field ($L_1$)} & NDCG&.508$\pm$.141  & .511$\pm$.128 & .534$\pm$.103 & .543$\pm$.084 & .544$\pm$.096 &.571$\pm$.089 & .578$\pm$.086 & .595$\pm$.089 & \textbf{.615$\pm$.084}  \\  
          ~&~& MRR &.556$\pm$.136  & .565$\pm$.105 & .610$\pm$.096 & .616$\pm$.076 & .622$\pm$.092&.649$\pm$.081 & .657$\pm$.078 & .675$\pm$.082 & \textbf{.702$\pm$.081}\\
    \cmidrule{2-12}
        ~ & \multirow{2}{*}{Paper--Field ($L_2$)} & NDCG &.318$\pm$.074 & .328$\pm$.046& .339$\pm$.049& .336$\pm$.062  & .342$\pm$.051&.350$\pm$.045 & .354$\pm$.046 & .358$\pm$.052&\textbf{.367$\pm$.048}\\
        ~&~& MRR &.322$\pm$.067 &.332$\pm$.052 &.348$\pm$.045 &.350$\pm$.053 &.358$\pm$.049 &.362$\pm$.057 &.369$\pm$.058&
        .371$\pm$.064& \textbf{.378$\pm$.071}\\
    \cmidrule{2-12}
        ~ & \multirow{2}{*}{\tabincell{c}{Paper--Venue }} & NDCG  &.302$\pm$.066
         & .313$\pm$.051 &.317$\pm$.057& .309$\pm$.071& .327$\pm$.062& .334$\pm$.058 &.341$\pm$.059  &.353$\pm$.064 &\textbf{.355$\pm$.062}  \\  ~&~& MRR &.194$\pm$.070 & .193$\pm$.047 &.196$\pm$.052 & .192$\pm$.059& .214$\pm$.067&.229$\pm$.061 &.233$\pm$.060 &.243$\pm$.048 & \textbf{.247$\pm$.061}\\
    \cmidrule{2-12}
        ~ & \multirow{2}{*}{\tabincell{c}{Author\\Disambiguation}} & NDCG &.738$\pm$.042 & .755$\pm$.048&.797$\pm$.044 &.803$\pm$.058 & .821$\pm$.056 & .835$\pm$.043&.841$\pm$.041 & .847$\pm$.043& \textbf{.852$\pm$.048}\\   ~&~& MRR &.612$\pm$.064&.619$\pm$.057 &.645$\pm$.063 & .649$\pm$.052 &.660$\pm$.049 &.668$\pm$.059 &.674$\pm$.058 &.683$\pm$.066 & \textbf{.688$\pm$.054}  \\
\bottomrule

\end{tabular}

\caption{Experimental results of different methods over the three datasets.} 
\label{tab:result} 
\end{table*}


\vpara{Input Features.}As we don't assume the feature of each node type belongs to the same distribution, we are free to use the most appropriate features to represent each type of nodes. 
For each paper, we use a pre-trained XLNet~\cite{xlnet, wolf2019transformers} to get the representation of each word in its title. 
We then average them weighted by each word's attention to get the title representation for each paper. 
The initial feature of each author is then simply an average of his/her published papers' representations. 
For the field, venue, and institute nodes, we use the metapath2vec model~\cite{dong2017metapath2vec} to train their node embeddings by reflecting the heterogeneous network structures.

The homogeneous GNN baselines assume the node features belong to the same distribution, while our feature extraction doesn't fulfill this assumption. 
To make a fair comparison, we add an adaptation layer between the input features and all used GNNs. 
This module simply conducts different linear projections for nodes of different  types. 
Such a procedure can be regarded to map heterogeneous data into the same distribution, which is also adopted in literature~\cite{DBLP:conf/kdd/ZhangSHSC19, DBLP:conf/www/WangJSWYCY19}. 

\vpara{Implementation Details.}
We use 256 as the hidden dimension throughout the neural networks for all baselines. For all multi-head attention-based methods, we set the head number as 8. 
All GNNs keep 3 layers so that the receptive fields of each network are exactly the same. 
All baselines are optimized via the AdamW optimizer~\cite{DBLP:conf/iclr/LoshchilovH19} with the Cosine Annealing Learning Rate Scheduler~\cite{DBLP:conf/iclr/LoshchilovH17}. For each model, we train it for 200 epochs and select the one with the lowest validation loss as the reported model. We use the default parameters used in GNN literature and donot tune hyper-parameters.

\subsection{Experimental Results}
We summarize the experimental results of the proposed model and baselines on three datasets in  Table ~\ref{tab:result}. 
All experiments for the four tasks are evaluated in terms of NDCG and MRR.

The results show that in terms of both metrics, the proposed \short\ significantly and consistently outperforms all baselines for all tasks on all datasets. 
Take, for example, the Paper--Field ($L_1$) classification task on OAG, \short\ achieves relative performance gains over baselines by 15--19\% in terms of NDCG and 18--21\% in terms of MRR (i.e., the performance gap divided by the baseline performance). 
When compared to HAN---the best baseline for most of the cases, the average relative NDCG improvements of \short\ on the CS, Med and OAG datasets are 11$\%$, 10$\%$ and 8$\%$, respectively. 

Overall, we observe that on average, \short\ outperforms GCN, GAT, RGCN, HetGNN, and HAN by 20\% for the four tasks on all three large-scale datasets.  
Moreover, \short\ has fewer parameters and comparable batch time than all the heterogeneous graph neural network baselines, including RGCN, HetGNN, and HAN. 
This suggests that by modeling heterogeneous edges according to their meta relation schema, we are able to have better generalization with fewer resource consumption.

\vpara{Ablation Study.}The core component in \short\ are the heterogeneous weight parameterization (Heter) and Relative Temporal Encoding (RTE).
To further analyze their effects, we conduct an ablation study by removing them from \short. 
Specifically, the model that removes heterogeneous weight parameterization, i.e., \short$_{-Heter}^{+RTE}$, drops 4\% of performance compared with the full model \short$_{+Heter}^{+RTE}$. 
By removing RTE (i.e., \short$_{+Heter}^{-RTE}$), the performance has a 2\% drop. 
The ablation study shows the significance of parameterizing with meta relations and using Relative Temporal Encoding.

In addition, we also try to implement a baseline that keeps a unique weight matrix for each relation. However, such a baseline contains too many parameters so that our experimental setting doesn't have enough GPU memory to optimize it. This also indicates that using the meta relation to parameterize weight matrices can achieve competitive performance with fewer resources.

\subsection{Case Study}

To further evaluate how  Relative Temporal Encoding (RTE) can help \short\ to capture graph dynamics, we conduct a case study showing the evolution of conference topic. 
We select 100 conferences in computer science with the highest citations, assign them three different timestamps, i.e., 2000, 2010 and 2020, and construct sub-graphs initialized by them. 
Using a trained HGT, we can get the representations for these conferences, with which we can calculate the euclidean distances between them. 
We select WWW, KDD, and NeurIPS as illustration. 
For each of them, we pick the top-5 most similar conferences (i.e., the one with the smallest euclidean distance) to show how the conference's topics evolve over time.

As shown in Table~\ref{tab:case}, these venues' relationships have changed from 2000 to 2020. 
For example, WWW in 2000 was more related to some database conferences, i.e., SIGMOD and VLDB, and some networking conferences, i.e., NSDI and GLOBECOM. 
However,  WWW in 2020 would become more related to some data mining and information retrieval conferences (KDD, SIGIR, and WSDM), in addition to SIGMOD and GLOBECOM. 
Also, KDD in 2000 was more related to traditional database and data mining venues, while in 2020 it will tend to correlate with a variety of topics, i.e. machine learning (NeurIPS), database (SIGMOD), Web (WWW), AI (AAAI), and NLP (EMNLP). 
Additionally, our \short\ model can capture the difference brought by new conferences. 
For example, NeurIPS in 2020 would relate with ICLR, which is a newly organized deep learning conference. 
This case study shows that the relative temporal encoding can help capture the temporal evolution of the heterogeneous academic graphs.

\begin{table}[t!]
\centering
\renewcommand\arraystretch{1.3}
\begin{tabular}{ccc} 
\toprule
Venue & Time & Top$-$5 Most Similar Venues \\
\midrule
\multirow{3}{*}{WWW} & 2000 & SIGMOD, VLDB, NSDI, GLOBECOM, SIGIR\\
~& 2010 & GLOBECOM, KDD, CIKM, SIGIR, SIGMOD\\
~& 2020 & KDD, GLOBECOM, SIGIR, WSDM, SIGMOD\\
\midrule
\multirow{3}{*}{KDD} & 2000 & SIGMOD, ICDE, ICDM, CIKM, VLDB\\
~& 2010 & ICDE, WWW, NeurIPS, SIGMOD, ICML\\
~& 2020 & NeurIPS, SIGMOD, WWW, AAAI, EMNLP\\
\midrule
\multirow{3}{*}{NeurIPS} & 2000 & ICCV, ICML, ECCV, AAAI, CVPR\\
~& 2010 & ICML, CVPR, ACL, KDD, AAAI\\
~& 2020 & ICML, CVPR, ICLR, ICCV, ACL\\
\bottomrule
\end{tabular}
\caption{Temporal Evolution of Conference Similarity.} 
\label{tab:case} 
\end{table}

\subsection{Visualize Meta Relation Attention}
To illustrate how the incorporated meta relation schema can benefit the heterogeneous message passing process, we pick the schema that has the largest attention value in each of the first two \short\ layers and plot the meta relation attention hierarchy tree in Figure~\ref{fig:meta}. 
For example, to calculate a paper's representation, 
$\langle$Paper, $is\_published\_at$, Venue, $is\_published\_at^{-1}$, Paper$\rangle$, 
$\langle$Paper, $has\_L_2\_field\_of$, Field, $has\_L_5\_field\_of^{-1}$, Paper$\rangle$, 
and $\langle$Institute, $is\_affiliated\_with^{-1}$, Author, $is\_first\_author\_of$, Paper$\rangle$
are the three most important meta relation sequences, which can be regarded as meta paths \textit{PVP, PFP,} and \textit{IAP}, respectively. 
Note that these meta paths and their importance are automatically learned from the data without manual design.  
Another example of calculating an author node's representation is shown on the right. 
Such visualization demonstrates that \model\ is capable of implicitly learning to construct important meta paths for specific downstream tasks, without manual customization.

\hide{
 \begin{figure}[t!]
    \centering
        \includegraphics[width=0.5\textwidth, trim = 30 0 50 0,clip]{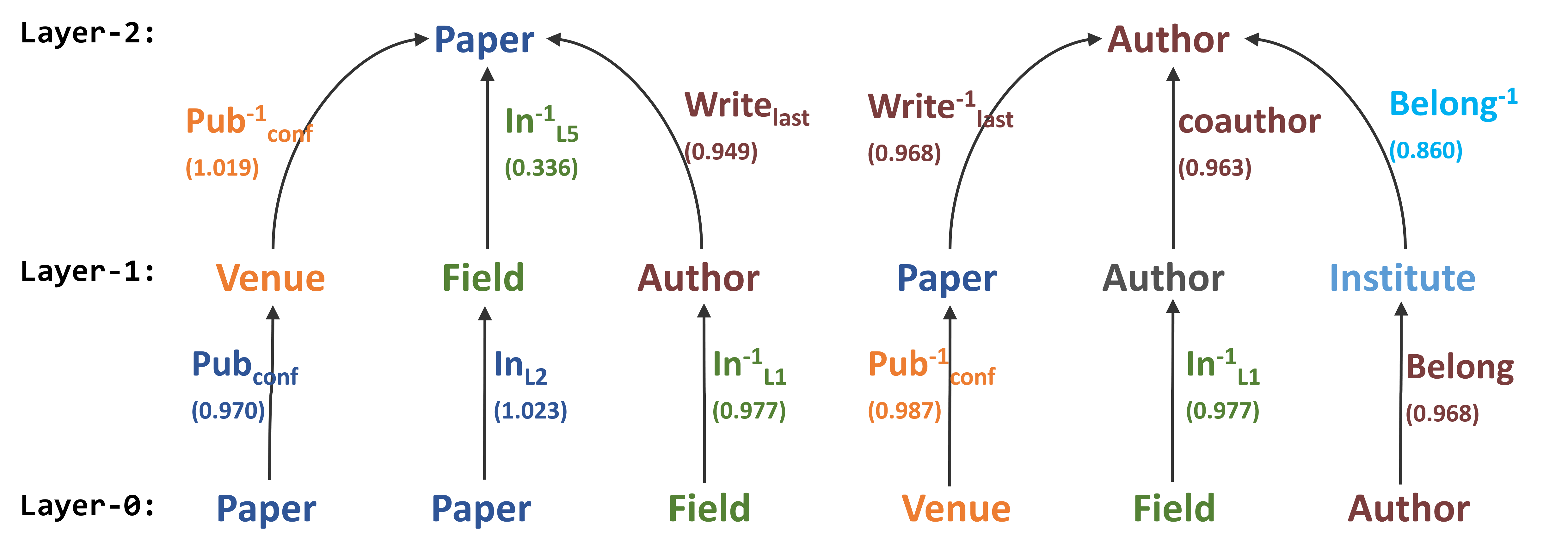}
    \caption{Hierarchy tree of learned meta relation attention.}
    \label{fig:meta}
\end{figure} 

}

 \begin{figure}[t!]
    \centering
        \includegraphics[width=0.46\textwidth, trim = 30 50 40 35,clip]{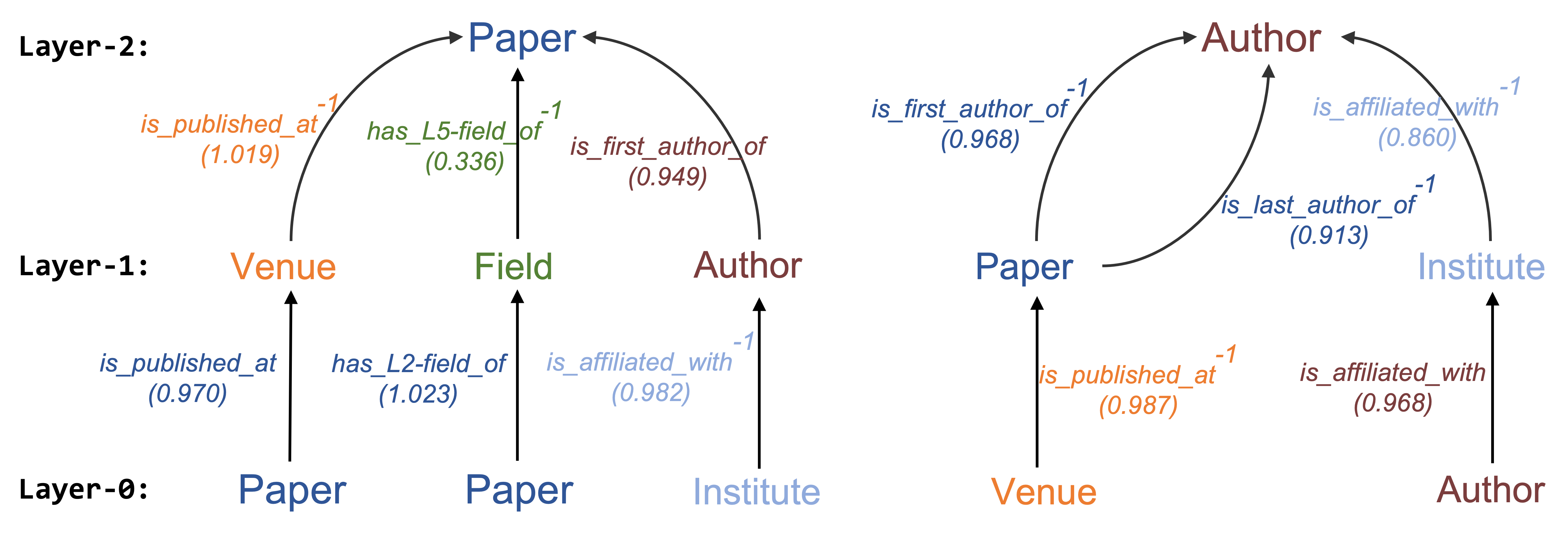}
    \caption{Hierarchy of the learned meta relation attention.}
    \label{fig:meta}
    \vspace{-0.3cm}
\end{figure}


%% file: section/conclusion.tex

In this paper, we propose the \model\ (\short) architecture for modeling Web-scale heterogeneous and dynamic graphs. 
To model heterogeneity, we use the meta relation $\langle \tau(s), \phi(e), \tau(t) \rangle$ to decompose the interaction and transform matrices, enabling the model to have the similar modeling capacity with fewer resources.
To capture graph dynamics, we present 
the relative temporal encoding  (RTE) technique to incorporate temporal information using limited computational resources. 
To conduct efficient and scalable training of \short\ on Web-scale data, we design the heterogeneous Mini-Batch graph sampling algorithm---\sampling. 
We conduct comprehensive experiments on the Open Academic Graph, and show that the proposed \short\ model can capture both heterogeneity and outperforms all the state-of-the-art GNN baselines on various downstream tasks.

In the future, we will explore whether \short\ is able to generate heterogeneous graphs, e.g., predict new papers and their titles, and whether we can pre-train \short\ to benefit tasks with scarce labels.

\vpara{Acknowledgements.} We would like to thank Xiaodong Liu for helpful discussions. This work is partially supported by NSF III-1705169, NSF CAREER Award 1741634, NSF\#1937599, Okawa Foundation Grant, and Amazon Research Award.